\pdfoutput=1 

\RequirePackage{pdf14}

\documentclass[10pt,journal,compsoc]{IEEEtran}
\newif\ifpeerreview

\peerreviewfalse
\usepackage[nocompress]{cite}
\usepackage{amsmath,amssymb,graphicx}

\usepackage{lipsum} 
\usepackage{amsbsy}
\usepackage[mathscr]{euscript}
\usepackage{amssymb}
\usepackage{amsmath}
\usepackage{stackrel}
\usepackage{changes}
\usepackage{nicefrac}
\usepackage{enumitem}

\usepackage{algpseudocode}
\usepackage[ruled]{algorithm2e} 

\SetAlFnt{\small}
\SetAlCapFnt{\small}
\SetAlCapNameFnt{\small}
\SetAlCapHSkip{0pt}
\IncMargin{-\parindent}

\usepackage{dblfloatfix}
\usepackage{float}

\usepackage{siunitx}
\usepackage{subcaption}

\usepackage{changes}

\usepackage[switch]{lineno}

\newcommand{\paperID}{0040}

\title{Stochastic Light Field Holography}

\author{Florian Schiffers, Praneeth Chakravarthula, Nathan 	Matsuda, Grace~Kuo, Ethan~Tseng,~Douglas~Lanman,~Felix Heide, Oliver Cossairt 
\IEEEcompsocitemizethanks{\IEEEcompsocthanksitem F. Schiffers is with the Computer
Science Department at Northwestern University
\IEEEcompsocthanksitem F. Schiffers, N. Matsuda, G. Kuo, D. Lanman and O. Cossairt are with Reality Labs Research, Meta
\IEEEcompsocthanksitem P. Chakravarthula, E. Tseng and F. Heide are with Princeton University
}
}

\begin{document}

\IEEEtitleabstractindextext{%
\begin{abstract}
The Visual Turing Test is the ultimate goal to evaluate the realism of holographic displays.
Previous studies have focused on addressing challenges such as limited étendue and image quality over a large focal volume, but they have not investigated the effect of pupil sampling on the viewing experience in full 3D holograms.
In this work, we tackle this problem with a novel hologram generation algorithm motivated by matching the projection operators of incoherent (\textit{Light Field}) and coherent (\textit{Wigner Function}) light transport.
To this end, we supervise hologram computation using synthesized photographs, which are rendered on-the-fly using Light Field refocusing from stochastically sampled pupil states during optimization.
The proposed method produces holograms with correct parallax and focus cues, which are important for passing the Visual Turing Test.
We validate that our approach compares favorably to state-of-the-art CGH algorithms that use Light Field and Focal Stack supervision.
Our experiments demonstrate that our algorithm improves the viewing experience when evaluated under a large variety of different pupil states.
\end{abstract}

}

\maketitle

\IEEEraisesectionheading{
  \section{Introduction}\label{sec:introduction}
}

\label{sec:intro}



\IEEEPARstart{I}{N} 1972, the Cartier jewelry store on 5th Avenue in New York City displayed an analog hologram of a hand-holding jewelry on their window storefront.
This hologram was so realistic that an elderly woman passing by attempted to attack the virtual arm floating in mid-air~\cite{walton2021characterizing}.
Large-format analog holograms are known for their incredible realism and ability to give the impression of a complete 3D picture frozen in time.
For static objects, full-color holography has been shown to provide realism on par with the visual inspection of the actual object.
As such, holography is often seen as the most likely approach to pass the Visual~Turing~Test~\cite{zhong2021reproducing}.
\begin{figure}[tb!]
    \centering
    \includegraphics[width=\linewidth]{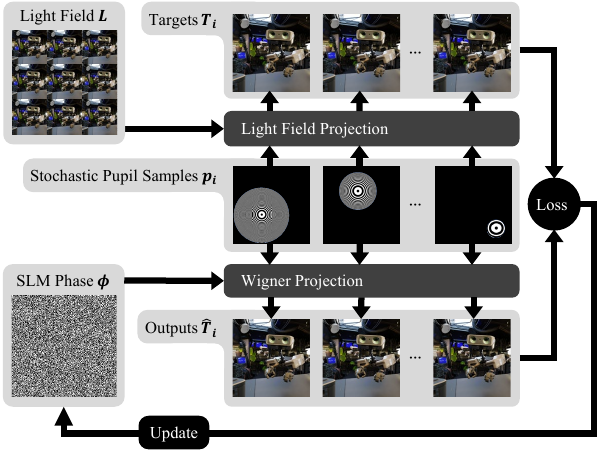}
    \caption{
    \textit{Stochastic Light Field Holography Algorithm:} We wish to infer an SLM phase image $\phi$ that produces a hologram with viewable intensity matching the projections rendered from a target Light Field $L$. To do this, a series of pupil samples with randomly generated position, diameter, and focus, $p_i$ are passed to two equivalent image formation models: a Light Field projection operator, which produces individual target images $T_i$, and a Wigner projection operator, which produces individual output estimates $\hat{T}_i$, for each of the pupil samples. These target and output estimates are compared using a photo-consistency loss function, which is in turn used to update the SLM phase image with a gradient descent step.
    }
    \label{fig:algorithm_overview}
\end{figure}
However, the dynamic display of high-quality digital holographic 3D imagery has proven challenging as a result of the limited system étendue.
Etendue is determined by the Space-Bandwidth-Product (SBP) of the modulator, which is equal to the product of the maximum diffraction angle supported by the modulator and its modulation area.
Most existing holographic systems rely on phase-only spatial light modulators (SLM), where this angle is limited by the pixel size -- today's fabrication techniques achieve 8$\mu$m pixel pitch for phase SLMs with no more than 4$^\circ$ of diffraction while limiting the eyebox to the size of the device. 
As a result, the available SBP mandates a \emph{trade-off between FOV and eyebox size that impacts the number of achievable angular views}.
%
%
The SBP required for holographic near-eye displays with a large eye box (roughly $10 \text{mm}$) and large FoV (~100 degree), is still so high that pupil steering using eye tracking might be necessary unless étendue-expansion approaches~\cite{kuo2020high, baek2021neural, monin2022exponentially,jo2022multi,monin2022analyzing} make significant progress.

\begin{figure*}
  \includegraphics[width=1.0\linewidth]{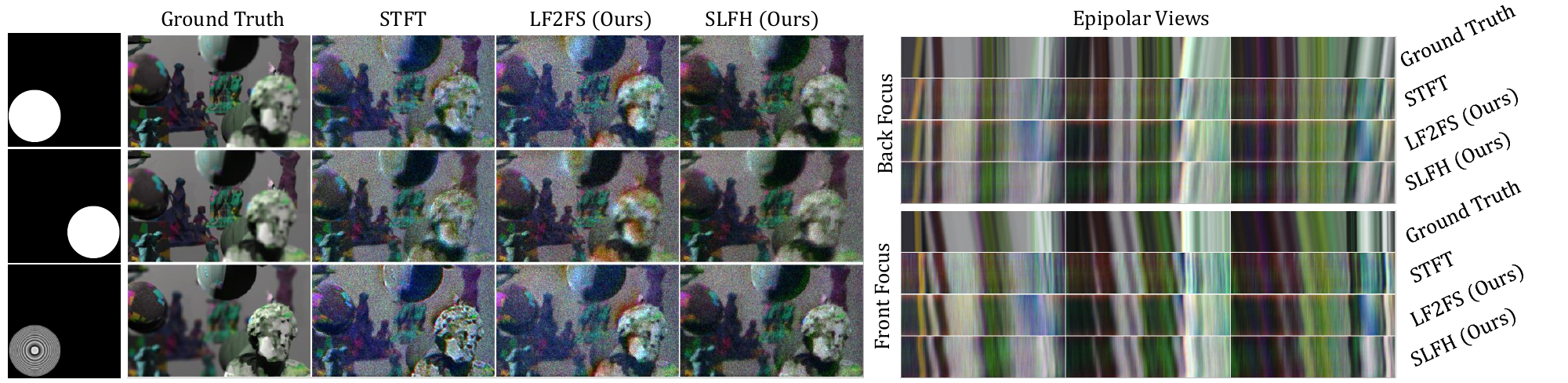}
  \caption{
  \emph{Stochastic Light Field Holography on experimental hardware.} 
  Recent advances in holographic displays achieve high image fidelity using smooth-phase holograms.
  This comes at the cost of a highly concentrated eye box limiting image formation to only the central view.
  For a more robust visual experience, holographic displays should provide good image quality for arbitrary pupil-states (\textit{diameter, location}, and \textit{accommodation}) within the entire available eye box.
  %
  %
  With \textit{Stochastic Light Field Holography} (SLFH), we propose a novel framework that ensures photo-consistency over the entire eye-box volume.
  We implement a novel \textit{Focal Stack} supervision algorithm (LF2FS) and show that SLFH is a generalization of state-of-the-art (SOTA) \textit{Focal Stack} and \textit{Short-Time-Fourier-Transform~(STFT)} supervision of CGH optimization algorithms, representing only a limited subset of possible pupil states.
  Consequently, our proposed method provides the best average image fidelity over the entire eyebox compared to SOTA.
  On the left, we show experimental captures from three example pupil states (\textit{Front/Back focus for accommodation}, \textit{left/right for parallax}). On the right, we show epipolar slices generated by capturing a 1D trajectory consisting of 31 small pupils (1/8th of eyebox) from left to right for both front and back focus.
  Our SLFH method produces the least artifacts over the full eyebox, while Focal Stack supervision produces strong color-fringing at occlusions and STFT-optimized holograms often show ringing artifacts and over-sharpened in focus images.
  %
  %
  %
  }
  \label{fig:teaser}
\end{figure*}

\begin{figure}[htbp]
    \centering
    \includegraphics[width=\linewidth]{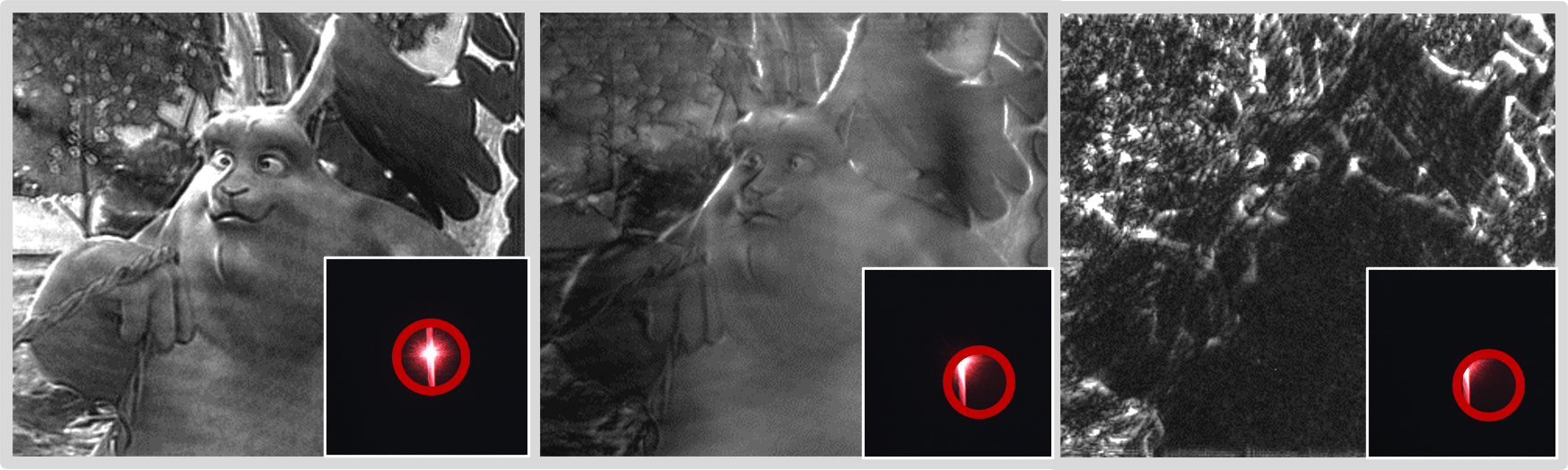}
    \vspace{-.2in}
    \caption{
    %
    \textit{The eyebox problem and Smooth-Phase CGH algorithms:}
    State-of-the-art holograms can achieve high image quality, but their spectral energy is highly concentrated leading to a small, effective eye-box. 
    Here, we show image degradation of smooth-phase holograms on our experimental setup once the eye-pupil (\textit{red inset}) does not sample the DC-term anymore.
    %
    \vspace{-4mm} 
    }
    \label{fig:dpac_comparison_eyebox}
\end{figure}

\textit{Problem Statement.}
Even with novel designs, achieving and utilizing a large enough SBP to fully realize the potential of holographic displays remains a challenge.
Specifically, a key question is how to compute holograms such that they provide an optimal image for \emph{any} possible viewpoint.
This problem is amplified for large SBP systems with large eyebox size as the pupil can move significantly within.
Pupil movement, including changes in location, size, and defocus, can significantly affect image quality on holographic displays~\cite{chakravarthula2022pupil}.
Accurately accounting for pupil movement is, as such, a critical challenge in the development of photo-realistic holographic displays, as it can greatly affect the realism and immersion of the viewing experience (see Fig.~\ref{fig:dpac_comparison_eyebox}). Only recently, researchers have investigated this issue for 2D holography~\cite{chakravarthula2022pupil} and showed only very preliminary 3D results using a two-plane representation with two objects. Hence, the computation of pupil-aware 3D holograms is an open problem that we address in this paper.

\textit{Proposed Solution.}
In this work, we address the challenge of computing high-quality holograms that accurately reproduce vision cues during pupil movement.
%
%
To do this, we propose a new Stochastic Light Field Holography (SLFH) algorithm that generates holograms using a loss function motivated by matching the photographic projection operators of incoherent and coherent light.
Our approach ensures optimal performance for a random assortment of pupil states and, unlike previous work, produces 3D imagery while also avoiding overfitting to specific viewing conditions.  
While previous methods use a loss function based on a set of static, prerendered images, we supervise our loss function by synthesizing novel views using Light Field refocusing~\cite{ng2005light} from arbitrary pupil states.
By sampling pupil states stochastically, we ensure, unlike existing work, that the optimized hologram provides correct parallax, defocus, and depth-of-field cues for all possible pupil states.
In order to have a fair comparison method with multi-plane approaches~\cite{choi2021neural,kavakli2022realistic}, we introduce a novel Focal Stack supervision algorithm (LF2FS) which ensures a parallax-consistent defocus allowing for a direct comparison to our main contribution, which is SLFH.

\begin{figure}
    \includegraphics[width=\linewidth]{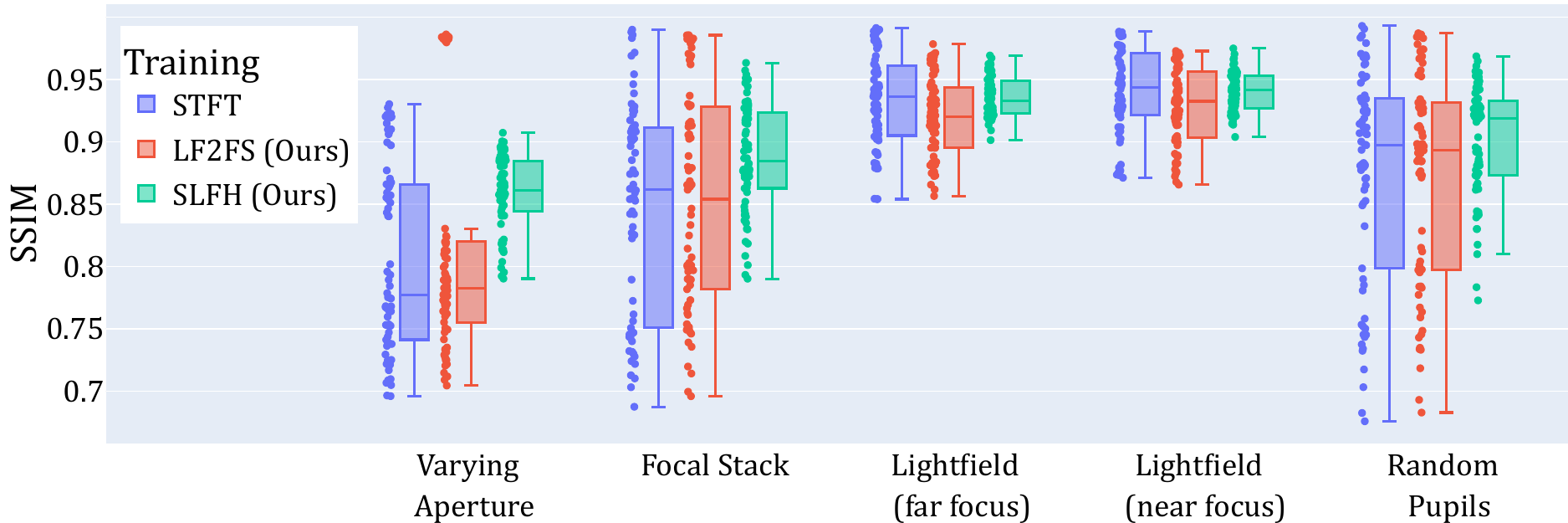}
  \caption{\textit{Statistical Evaluation:} Statistical performance evaluation of SSIM percpetual similarity metric over the Deep Spaces~\cite{xiao2018deepfocus} light field dataset. Holograms are trained using STFT, Focal Stack supervision (LF2FS), and Stochastic Light Field Holography (SLFH).
  Statistics are plotted for various aperture types: Varying Aperture (focus and position are fixed), Focal Stack (aperture diameter and position are fixed), Light Field (aperture diameter and focus are fixed), and random pupils.
  Each dot represents the average SSIM of all evaluated pupil-positions for a specific aperture type for one specific light field within the Deep Spaces~\cite{xiao2018deepfocus} dataset.
  Hence, each dot corresponds to the average performance for a specific scene.
  Our method always produces the best worse case performance, and produces a better mean and variance for random pupils, which means that our algorithm performs better for a wider variety of near-eye viewing conditions.
      \vspace{-4mm} 
}
  \label{fig:results_statistics_deepfocus}
\end{figure}

\textit{Contributions.} Our software codebase for stochastic pupil sampling  \textit{will be made publicly available after publication} of this manuscript. Our paper makes the following contributions:
\begin{itemize}[leftmargin=*]
    \item We are the \emph{first} to \emph{study} the image quality of 3D-CGH algorithms for all possible pupil states within the available eye-box volume.
    \item We are the \emph{first} to \emph{introduce} on-the-fly Light Field refocusing \cite{ng2005fourier} into the holographic generation process.
    We \emph{propose} a new CGH-algorithm that reproduces parallax and accommodation for arbitrary pupil states by coupling the projection operators for ray-space \emph{(Light Field)} and wave-optics \emph{(Wigner Distribution)}. We use this to implement both a novel Focal Stack supervision algorithm (LF2FS) and our proposed SLFH method.
    \item We \emph{validate} the method in simulation and demonstrate that, while previous techniques overfit to specific pupil states used during training, our method produces the most favorable display for arbitrary viewing conditions (see Fig.~\ref{fig:results_statistics_deepfocus}). 
    \item We \emph{assess} the method with an experimental prototype and demonstrate higher quality parallax information with fewer artifacts than state-of-the-art Light Field supervision CGH optimization algorithms (see Figs.~\ref{fig:teaser},~\ref{fig:results:focal_stack_experiments}, and Supplemental Material).
    %
\end{itemize}

\section{Related Work}
\label{sec:related}

%
%
Holography for displays relies on diffraction and interference of light to generate images.
Based on the diffracted field, a hologram can be classified as a far-field Fourier hologram or a near-field Fresnel hologram.
Using phase-only SLMs requires computing phase-only holograms that are capable of producing the diffraction field that can closely mimic the target image. However, the underlying phase retrieval problem is generally ill-posed and non-convex.
 Though introduced for Fourier phase retrieval, early methods such as error reduction using iterative optimization~\cite{lesem1969kinoform,gerchberg1972practical} and hybrid input-output (HIO) methods~\cite{fienup1982phase, bauschke2003hybrid} are applicable for both Fourier and Fresnel holograms.
 Researchers have also explored phase-retrieval methods using first-order nonlinear optimization \cite{lane1991phase}, alternative direction methods for phase retrieval~\cite{wen2012alternating,marchesini2016alternating}, nonconvex optimization~\cite{zhang20173d}, and methods overcoming the nonconvex nature of the phase retrieval problem by lifting, i.e., relaxation, to a semidefinite~\cite{candes2013phaselift} or linear program~\cite{goldstein2018phasemax,bahmaniPhaseMax}. Several works~\cite{peng2020neural,chakravarthula2019wirtinger} have explored optimization approaches using first-order gradient descent methods to solve for holograms with flexible loss functions.

\textit{Focal Stack and RGBD Algorithms.}
Instead of computing the wave propagation for millions of points, a 3D object can be represented as a stack of intensity layers~\cite{zhang2016layered, zhao2015accurate,choi2021neural,eybposh2020deepcgh}.  
Wave propagation methods such as the inverse Fresnel transform or angular spectrum propagation are typically used for propagating the waves from several layers of the 3D scene towards the SLM plane, where they are interfered with to produce a complex hologram~\cite{kuo2020high,peng2020neural,zhang20173d,makowski2007iterative}.
Although this approach can be implemented efficiently, it cannot support continuous focus cues and accurate occlusion due to discrete plane sampling. 
Recent work~\cite{kim2022accommodative} suggests that holograms based on smooth-phase profiles~\cite{yoo2021optimization} cannot support accommodation; the ultimate promise of 3d holography.
An additional discussion on smooth vs random phase holograms is found in the supplementary.

A further approximation to the layer-based methods is to determine the focal depth of the user, i.e., distance of the object to which the user fixates via an eyetracker and adjusting the focal plane of the 2D holographic projection to match the user focal distance~\cite{maimone2017holographic}.
While emulating a 3D scene by adaptively shifting a 2D holographic projection in space is computationally efficient, operating in a varifocal mode under-utilizes the capabilities of a holographic display.
Moreover, achieving natural focus cues and physically accurate occlusion effects still remains a challenge. 
To compare against SOTA Focal Stack methods, \textit{we introduce a novel Focal Stack supervision algorithm (LF2FS)} that generates Focal Stacks on the fly from Light Fields according to physically accurate pupil states. 

\textit{Light Field Algorithms.}
\label{sec:RelatedWork:LightFieldAlgorithms}
To support occlusion and depth-dependent effects, a Light Field can be encoded into a hologram partitioned spatially into elementary hologram patches, called ``hogels'' \cite{zhang2015fully}, similar to elementary images in a Light Field.
These hogels produce local ray distributions that reconstruct multiple (Light Field) views~\cite{lucente1995rendering,yamaguchi1993phase,smithwick2010interactive}.
Such holograms which encode a Light Field are called ``holographic stereograms''.
Conventional stereograms, where hogels are out of phase with each other, suffer from a lack of focus cues and limited depth of field~\cite{lucente1995rendering}. 
%
%
To keep the hogels of a holographic stereogram in phase throughout the hologram, researchers have introduced an additional phase factor to calculate what is called a phase-added stereogram (PAS)~\cite{yamaguchi1993phase}. 
%
%
However, akin to a microlens array-based Light Field display~\cite{lanman2013near}, stereograms suffer from the fundamental spatio-angular resolution trade-off:
A larger hogel size leads to a decreased spatial resolution.
This fundamental limitation does not allow for high spatial resolution holographic stereogram projections.
Recent methods have attempted to overcome this trade-off~\cite{padmanaban2019holographic,blinder2018accelerated,park2019non} via Short-Time Fourier Transform (STFT) inversion.
However, these methods do not match the image quality achieved for 2D holograms~\cite{chakravarthula2020learned,peng2020neural} and suffer from artifacts around object discontinuities due to suboptimal inversion of the STFT.
More importantly, these algorithms do not incorporate pupil models for viewing the hologram, and therefore overfit to viewing conditions where the eyebox and pupil coincide. 
%
%
%
%
%
In Sec.~\ref{section:photographic_projection_operators} will further discuss the Wigner function and how it relates to holographic displays.
We want to stress that we are not the first to consider the Wigner-function for holographic displays~\cite{shi2017near,park2019non,min2023wigner}.
However, these methods use the Wigner-function to establish a closed-form solution similar to~\cite{padmanaban2019holographic}, which is different from ours, which uses an iterative approach to recover the displayed phase-modulation.

\textit{Learning-Based Holography.}
Neural networks and deep learning approaches have recently been proposed as tools for optical design and holographic phase retrieval.
Holographic microscopy has been tackled by solving phase retrieval problems using neural networks~\cite{rivenson2018phase,eybposh2020deepcgh}.
In a similar fashion, neural networks have been investigated for learning holographic wave propagation from a large training dataset.
For example, Horisaki et al.~\cite{horisaki2018deep} trained a U-net on a pair of SLM phase and intensity patterns, and predicted SLM phase patterns during inference.
Recently, Eybposh et al.~\cite{eybposh2020deepcgh} proposed an unsupervised training strategy and predicted the SLM phase patterns in real time that produced 2D and 3D holographic projections.
Peng et al.~\cite{peng2020neural} and Chakravarthula et al.~\cite{chakravarthula2020learned} have recently demonstrated camera-in-the-loop (CITL) calibration of hardware using neural networks and high-fidelity holographic images on prototype displays.
Shi et al.~\cite{shi2021towards} have demonstrated high-resolution real-time holography with a light weight neural phase-retrieval network that may be suitable for inference on mobile hardware in the future.

\textit{Pupil and Parallax-aware Holography.}
Very recently, researchers have investigated the effects of pupil sampling on holographic setups.
Chakravarthula et al. \cite{chakravarthula2022pupil}  work is the most similar work related to the investigation in this paper.
Chakravarthula et al. propose a new cost function which enforces a uniform energy distribution over the eye-box by stochastically sampling different pupil positions. However, their approach is limited to 2D-imagery and very preliminary multi-plane images, and, as a result, they did not account for parallax and focus-cues effects.
Methods that operate on RGB-D scenes~\cite{peng2020neural,choi2021neural,shi2021towards} are capable of finding correct accommodation cues by using multi-plane or Focal Stack representations, however, are not able to account for correct parallax cues when the pupil is shifting. Motivated by this, recent work has proposed holographic generation algorithms using Light Field supervision \cite{choi2022time,padmanaban2019holographic,chakravarthula2022hogel}, making use of the Short Time Fourier Transform (STFT) and allowing a bidirectional transform between holograms and Light Fields as proposed by~\cite{ziegler2007bidirectional}. All of these existing methods have in common that they do not allow for pupil sampling and accurate parallax cues at the same time.

\section{Photographic Projection Operators}
\label{sec:wigner-analysis}
\label{section:photographic_projection_operators}
In this section, we establish a connection between the coherent \textit{(wave-optics)} and incoherent \textit{(ray-space)} projection operators.
We first introduce Light Fields, their coherent equivalent using the Wigner-formalism, and their resemblance to STFT approaches used in current literature. 

\begin{figure}
    \centering
    \includegraphics[width=\linewidth]{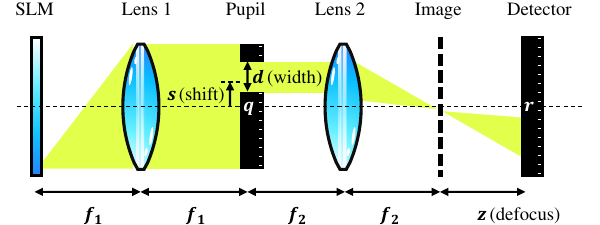}
    \caption{\textit{Simplified diagram of the proposed setup:} An SLM (illumination path not shown) diffracts a collimated beam according to an optimized pattern. The modulated beam passes through a 4-f system, with a pupil having programmable shift $s$ and diameter $d$. The relay system produces an image of the SLM offset from the detector by a defocus distance~$z$.
    \vspace{-5mm} 
}
    \label{fig:eye_figure_geometry}
\end{figure}
\vspace{-2mm}
\subsection{Eye Pupil Model}
\setlength{\belowdisplayskip}{4pt} \setlength{\belowdisplayshortskip}{4pt}
\setlength{\abovedisplayskip}{4pt} \setlength{\abovedisplayshortskip}{4pt}
The primary contribution of our work is to introduce a simplified eye/pupil model into the hologram forward model for optimization of photorealistic 3D imagery over a variety of viewing conditions. We approximate a near-eye holographic display system as a 4f relay where an eyepiece images an SLM illuminated by monochormatic light through a pupil, followed by an eye lens which images onto a detector (see Fig.~\ref{fig:eye_figure_geometry}). We represent 2D coordinates in the detector plane as $\mathbf{r}=(x,y)$, and pupil plane as $\mathbf{q}=(u,v)$. The eye is assumed to be focused at infinity in the rest state so that converting from pupil to angular coordinates is given by the relation $\mathbf{q} = \tan(\mathbf{\theta})/f$, where $f$ is the focal length of the eye model or objective lens ($f_2$ in Fig.~\ref{fig:eye_figure_geometry}). For simplicity, we parameterize the pupil state $p$ with a set of 4 parameters, that is
\begin{equation}
    p = \{ \mathbf{s}, z, d \},
\end{equation}

\noindent corresponding to focus distance $(z)$, 2D-pupil shift $\mathbf{s}=(x,y)$, and aperture diameter $(d)$. We define the pupil aperture function as a circular function that focuses at infinity in the rest state
\begin{equation}
    A(\mathbf{q};p) = 
\begin{cases}
    1,              & \text{if } |\mathbf{q}-\mathbf{s}| < d\\
    0,              & \text{otherwise.}
\end{cases}
\end{equation}

\vspace{-2mm}
\subsection{Light Fields}
Light Fields represent the flow of radiance in a scene in every direction and every point in space.
%
%
This extra information can be used to generate synthetic 2D images from different viewpoints and simulate optical effects like defocus and lens aberrations. 
The photography operator is used in digital refocusing~\cite{ng2005fourier} to render images at different depths from the Light Field under a virtual aperture function.
%
%
Specifically, the operator $\mathbf{P_{lf}}$ describes the transformation of a 4D Light Field $L(\mathbf{r},\mathbf{q})$ into a synthetic photograph focused through an aperture function $A$  with a depth defined by $\alpha$. We parameterize the Light Field such that $\mathbf{r}$ and $\mathbf{q}$ represent the spatial coordinates of the Light Field on detector and pupil planes, respectively.
The photographic projection operator can then be expressed as the following projection from 4D to 2D
\begin{equation}
\mathbf{P}_{lf}\left[L ; p \right](\mathbf{r}) = \int A(\mathbf{q};p) \cdot L(\mathbf{q} - \frac{\mathbf{q}-\mathbf{r}}{\alpha}, \mathbf{q}) \, d\mathbf{q},
\label{eq:lf_refocusing_formula_incoherent}
\end{equation}
Equation~\ref{eq:lf_refocusing_formula_incoherent} can be thought of as shearing the 4D space, multiplication with the aperture function and a subsequent projection down to 2D.
Note that the effects of changing focus distance $z$ are incorporated into the defocus parameter $\alpha = f/(z+f)$. 

\subsection{Wigner Function}
\label{subsec:wignerfunction}
The projection operator for Light Fields in Eq.~\ref{eq:lf_refocusing_formula_incoherent} works only for incoherent light as it cannot account for interference effects.
To overcome these limitations, \cite{zhang2009wigner,cuypers2011validity} used the Wigner Distribution-Function (WDF) to establish the connection to Light Fields. 
While it is computationally intractable to use Wigner functions efficiently for large holograms, it provides a powerful tool for understanding optical systems and their limitations.
%
%
%
To illustrate this, let us consider a monochromatic optical field $u( \mathbf{r} )$.
%
The correlation between two points on the field $\mathbf{r}$ and $\mathbf{r'}$ is given by the mutual intensity function, which operates on a 2D field and then produces a 4D quantity as
\begin{equation}
\text{J}[u(\mathbf{r})]\left( \mathbf{r}, \mathbf{r'} \right)=\left\langle u\left( \mathbf{r}+\frac{\mathbf{r'} }{2}\right) u^*\left(\mathbf{r}-\frac{\mathbf{r'} }{2}\right)\right\rangle,
\end{equation}
where $\langle \cdot \rangle$ denotes a time-average.
%
The WDF is then defined as the Fourier transform of the mutual intensity along $\mathbf{r'}$
\begin{equation}
\text{W}[u(\mathbf{r})]{ \left( \mathbf{r}, \mathbf{q'} \right) }=\int \text{J}\left( \mathbf{r}, \mathbf{r'}  \right) \exp \left(-2 \pi i \mathbf{r'} \cdot \mathbf{q'} \right) \mathrm{d} \mathbf{r'} .
\label{eq:wigner_projection_continuous}
\end{equation}
\noindent The spatial frequency coordinates $\mathbf{q'}$ of the WDF can be converted to spatial pupil coordinates $\mathbf{q}$ via the relation
\begin{equation}
    W(\mathbf{r},\mathbf{q}) = W(\mathbf{r},\lambda f \mathbf{q'}) .
\end{equation}
%
%

\textit{Wigner Projection Operator.}
%
%
%
We next consider the WDF $W(\mathbf{r},\mathbf{q})$ defined at the plane of a phase-only SLM with displayed phase pattern $\phi(\mathbf{r})$, producing a diffracted field $u(\mathbf{r}) = \exp{(j\phi(\mathbf{r}))}$.  
Defining the WDF of the aperture function $A(\mathbf{q} ; p)$ to be $W_A(\mathbf{r},\mathbf{q})$, the WDF that is imaged onto the retina of the eye (assuming unit magnification for simplicity) is 
\begin{equation}
    W_o[u(\mathbf{r})](\mathbf{r}, \mathbf{q})=\int W_A\left(\mathbf{r}, \mathbf{q}_i \right) W[u(\mathbf{r})]\left(\mathbf{r}, \mathbf{q}-\mathbf{q}_i\right)  \mathrm{d} \mathbf{q}_i \enspace
\end{equation}
The effect of a focus shift $z$ in Wigner-space can be represented as a shear~\cite{bastiaans2009wigner} given by 
\begin{align}    
    W_z\left(\mathbf{r}, \mathbf{q}\right) =    W_o \left( \mathbf{r} - \lambda  z  \mathbf{q'} , \mathbf{q} \right) = W_o \left( \mathbf{r} - \frac{z}{f} \mathbf{q} , \mathbf{q} \right) \enspace,
\end{align}
%
%
which results in the Wigner projection operator 
%
%
\begin{multline}
\mathbf{P}_{WF} \left [ u(\mathbf{r}) ; p \right]( \mathbf{r} ) = \\
\iint W_A(\mathbf{r},\mathbf{q}-\mathbf{q_i}) \cdot W[u(\mathbf{r})] \left( \mathbf{q} - \frac{\mathbf{q} - \mathbf{r}}{\alpha}, \mathbf{q} \right) d\mathbf{q}d\mathbf{q_i} . 
\label{eq:lf_refocusing_formula_coherent}
\end{multline}

%
%
%

\textit{Relationship between Wigner and Light Field Projection.}
%
Comparing Eq.~\ref{eq:lf_refocusing_formula_coherent} with Eq.~\ref{eq:lf_refocusing_formula_incoherent}, we observe that the projection geometry is identical while the $\text{WDF}$ projection allows us to incorporate interference effects from the eye pupil that naturally arise when viewing a holographic display illuminated with coherent light. Closer inspection reveals that the Wigner projection operator is a generalization of Light Field projection. When the retina detector pixel size $\Delta$ is much larger than the diffraction airy disc size $\Delta >> \lambda \cdot f/d$, then the effect of diffraction from the pupil aperture on the perceived image is negligible.
In this case, the WDF for the aperture is approximately independent of spatial coordinates $\mathbf{r}$ so that $  W_A(\mathbf{r}, \mathbf{q}) \approx A(\mathbf{q};p) \cdot W\left(\mathbf{r}, \mathbf{q}\right)$.
Hence, the projection operators for Eq.~\ref{eq:lf_refocusing_formula_coherent} and Eq.~\ref{eq:lf_refocusing_formula_incoherent} are identical within a scale factor.
\textit{Wigner Projection with Wave Optics.}
The similarity between photographic projection operators for Light Fields and WDFs provides an intuition for how to optimize holograms that produce photorealistic imagery over a variety of viewing conditions.
However, while the WDF is an elegant way to describe coherent light transport, it is impractical to compute on large fields~\cite{hamann2018time}.  
Instead, we implement the projection operator of Eq.~\ref{eq:lf_refocusing_formula_coherent} using wavefront propagators via stochastic pupil-sampling as described in Sec~\ref{subsec:implementation_projection_operator}.

\subsection{Bidirectional Hologram Light Field Transform}
As discussed in Sec.~\ref{sec:RelatedWork:LightFieldAlgorithms}, a successful branch of LF-CGH algorithms employs STFT-inversion to move between LFs and hologram-domains \cite{ziegler2007bidirectional}.
The Wigner projections (Sec.~\ref{subsec:wignerfunction}) have striking similarities with the STFT, as we will show in the following.
The STFT of a signal $u(\mathbf{r})$ is defined as the convolution with a window function $w(\mathbf{r})$ such that
\begin{equation}
\begin{aligned}
\text{STFT} [u( \mathbf{r} )]\left(\mathbf{r} , \mathbf{q} \right) &=
 \int u\left(\mathbf{r}^{\prime} \right) w\left( \mathbf{r}^{\prime} - \mathbf{r} \right) e^{-j 2 \pi \mathbf{q} \cdot \mathbf{r}^{\prime}} d \mathbf{r}^{\prime}  \\
 &= \mathcal{F}^{-1}\left\{ U\left(\mathbf{q'} \right) \cdot W\left(\mathbf{q'} - \mathbf{q}\right) \right\} ,
\end{aligned}
\label{eq:STFT1}
\end{equation}

\noindent where $U\left(\mathbf{q'} \right)$ and $W\left(\mathbf{q'} \right)$ are Fourier transforms of the field $u\left(\mathbf{r} \right)$ and window function $w\left(\mathbf{r} \right)$, respectively. The relationship between the STFT of the field $u(\mathbf{r})$ and the target lightfield is then
\begin{equation}
L\left(\mathbf{r}, \mathbf{q} \right)  =\left| \text{STFT} \left[ u(\mathbf{r}) \right]\left( \mathbf{r}, \mathbf{q} \right)\right|^2   .
\label{eq:STFT2}
\end{equation}
Together, Eqns. ~\ref{eq:STFT1} and ~\ref{eq:STFT2} express the relationship between STFT-based CGH algorithms and the Wigner projection operator implemented using wave optics as detailed in Sec.~\ref{sec:implementation}. Implicitly, the STFT computes an image of the hologram as viewed with a pupil function $W\left(\mathbf{q'} \right)$, which is the Fourier Transform of the STFT window function $w\left(\mathbf{r} \right)$. Typically a Hanning window or similar function is used to suppress ringing, and a grid of pupil positions is computed by applying a 2D FFT to the windowed field $u\left(\mathbf{r}^{\prime} \right) w\left( \mathbf{r}^{\prime} - \mathbf{r} \right)$. 

The method we propose in this work can be thought of as a generalization of STFT-based methods, with a few important differences. First, while the STFT method uses fixed pupils for hologram generation that provide photo-consistency with lightfield views, our method ensures that arbitrary pupil functions applied to the holographic display are consistent with Light Field projection using the same pupil. Second, our method relies on the relationship between spatial frequency coordinates and angular frequency coordinates of the field, which is necessary to produce accurate color reconstructions with multiple wavelengths. Lastly, the proposed method decouples the number of pupil positions from the forward model so that less memory can be used to compute larger holograms.            



\subsection{Eyebox, Pupils and Speckle}
This paper focuses on the problem of developing optimization algorithms for holograms that produce photorealistic 3D imagery over a variety of random pupil states. A necessary requirement for these algorithms is that they produce an eyebox with a relatively smooth distribution of intensity so that there is not a drastic change in perceived brightness as the pupil state changes during hologram viewing. In order for a hologram to produce a uniform eyebox, it must also produce a diffracted field with a relatively even distribution of angular frequencies, which means that the CGH will typically have highly randomized phase. This has implications on the type of CGH algorithm used, as well as the quality of imagery displayed since random phase holograms naturally introduce speckle.   

Many recent studies on holography do not address the issue of eyebox uniformity, and it is often assumed that the eye box is solely determined by the system etendue, which defines the maximum diffraction angle given by the SLM-pixel pitch. 
However, for most near-field hologram algorithms that have recently reported great image quality~\cite{peng2020neural,shi2021towards}, the light distribution in the eye-box is heavily concentrated around the DC term.
This concentration of spectral energy leads to a significant reduction in the "effective eye-box size".
%
This problem arises naturally in many existing holographic generation algorithms.
Double Phase (DPAC) encoding provides a direct method for encoding a complex field onto a phase only SLM, but only works well with smooth phase holograms that contain a Fourier spectrum highly concentrated around the optical axis (DC angular frequency).
Likewise, similar characteristics are observed when using holographic generation based on gradient-descent style algorithms that are initialized with smooth phase (see Fig.~\ref{fig:dpac_comparison_eyebox}).


Speckles are interference effects that occur when a diffracted field is highly random compared to its effective aperture.
The speckle size on the retina of a pupil sampled hologram is inversely proportional to the pupil diameter $d$~\cite{goodman2005introduction}.
In this work, we focus on CGHs which are viewed with pupil diameters much smaller than the eyebox ($d \in [w_{eyebox}/12, w_{eyebox}]$), which can introduce significant speckle noise in simulated and captured imagery relative to sampling the full eyebox.
To improve image quality and reduce the effect of speckle {we introduce incoherent averaging in the form of temporal averaging of 8 frames and spatial averaging of 2 $\times$ 2 blocks of pixels.}
For fair comparisons, we apply the same incoherent averaging to all algorithms used in this paper.

\section{Stochastic Light Field Holography}
\label{sec:3dpupil}

Building on the projection operators from the previous section, we now introduce the proposed method. We first introduce the forward model that describes the mapping from a displayed phase-modulation $\phi$ on the SLM to the complex wavefront $u$, assuming that the image is captured under an arbitrary pupil state. Then, we will introduce the proposed novel CGH-algorithm using Light Field supervision (see Fig.~\ref{fig:algorithm_overview}).

\subsection{Enforcing Photo-Consistency}

The proposed method enforces photo-consistency between image formation of a known Light Field $L(\mathbf{r},\mathbf{q})$ and a unknown Wigner distribution $W[u(\mathbf{r})](\mathbf{r},\mathbf{q})$, where $u(\mathbf{r}) = \exp{(j\phi(\mathbf{r}))}$ is the complex field formed by the holographic display with phase-only slm pattern $\phi(\mathbf{r})$.
%
We implement projection operators for both Light Field $P_{lf}\left[\cdot  \right]$ (see Eq.~\ref{eq:lf_refocusing_formula_incoherent}) and the Wigner Distribution $P_{wd} \left[ \cdot  \right]$ (see Eq.~\ref{eq:wigner_projection_continuous}) such that they match the geometry of our holographic display.
The projections of the known Light Field produce a set of target photographs given the viewing parameters 
\begin{equation}
    T_i( \mathbf{r} ) = P_{lf}[L, p_i] .
\end{equation}
We then generate the corresponding Wigner projections corresponding to observations of the diffracted optical field with the same viewing parameters 
\begin{equation}
    \hat{T}_i(\mathbf{r}, \phi(\mathbf{r})) = P_{wd}[\exp{(j\phi(\mathbf{r}))}, p_i] .
\end{equation}
The optimization objective is then to find the SLM pattern $ \phi(\mathbf{r})$ that solves
\begin{equation}
    \phi(x)^* = \text{argmin}_{\phi} \sum_i \mathscr{L}( T_i(\mathbf{r}),  \hat{T}_i(\mathbf{r}, \phi(\mathbf{r}))) 
\end{equation}
\begin{equation}
\text { given } p_i=\left\{
\begin{array}{l}
x_i \sim \mathcal{U}\left[-r_{\max }, r_{\max } \right] \enspace \enspace \enspace
z_i \sim \mathcal{U}\left[z_{\min },z_{\max }\right]
 \\
 y_i \sim \mathcal{U}\left[-r_{\max }, r_{\max } \right] \enspace \enspace \enspace
d_i \sim \mathcal{U}\left[r_{\min }, r_{\max }\right]
\end{array}
\right.
\end{equation}
where $\mathcal{U}$ is the uniform distribution and $\mathscr{L}$ is an arbitrary loss-function ($L_2$ in our case).
%
%
\vspace{-2mm}
\subsection{Implementation of Projection Operators}
\label{subsec:implementation_projection_operator}
\textit{Wigner-Projection using Wave Optics.}
%
While the Wigner projection operator of Eq.~\ref{eq:wigner_projection_continuous} is useful to build an intuition for how to optimize a hologram that is radiometrically consistent with a target lightfield, \emph{computing the full WDF is impractical} because the WDF is a 4D function which results in quartic memory growth. Instead, we use a Fourier optics forward model to stochastically sample pupil positions, resulting in the simplified 2D projection operator that does not require storing a 4D WDF, that is
\begin{equation}
    P_{wd}[u(\mathbf{r}), p](\mathbf{r}) = \int U(\mathbf{q}) K(\mathbf{q}, p) e^{-j 2 \pi \mathbf{q} \cdot \mathbf{r}} d \mathbf{r}.
    \label{eqn:2D_wigner_prop}
\end{equation}
Here, $U$ is the frequency representation of the phase-only SLM-modulation 
\begin{equation}
    U\left( \mathbf{q} \right)= \mathcal{F}\left\{u(\mathbf{r}\right)\}
\end{equation}
and the aperture function $K$ incorporates the effects of the pupil parameters $p$ through the the aperture function $A$, and the propagation kernel $\mathcal{H}$ as
\begin{equation}
    K(\mathbf{q}, p) = A\left(\mathbf{q} - \frac{\mathbf{s}}{\lambda f} , \frac{d}{\lambda f}\right)\mathcal{H}\left( \mathbf{q}; z\right).
\end{equation}
The angular spectrum propagation kernel $\mathcal{H}$ defined in the frequency domain is given by
\begin{equation}
    \mathcal{H}\left( \mathbf{q} ; z\right)= \begin{cases}e^{i \frac{2 \pi}{\lambda}} \sqrt{1- ||\lambda \mathbf{q} ||^2  z}, & \text { if } \sqrt{||\mathbf{q}||^{2}}<\frac{1}{\lambda}, \\
0 & \text { otherwise. }\end{cases}
\label{eq:aperture_function}
\end{equation}

\textit{Relationship to Focal Stack and STFT Supervision.}
The propagation kernel expressed by Eqn.~\ref{eqn:2D_wigner_prop} is a generalization of both Focal Stack \cite{choi2021neural,kavakli2022realistic} and STFT ~\cite{choi2022time} supervision applied to hologram optimization. For focal stack supervision, the pupil parameters $p=\left({\mathbf{s},d,z}\right)$ are fixed so that pupil shift $\mathbf{s}=0$, the aperture diameter $d$ is equal to the eyebox size, and the depth is discretized into $N$-layers over $z \in [z_{min},z_{max}]$. As discussed in Sec.~\ref{subsec:wignerfunction}, STFT supervision closely resembles Eq.~\ref{eqn:2D_wigner_prop}, but the kernel $K(\mathbf{q},p)$ is equal to the Fourier Transform of the STFT window $W(\mathbf{q})$, and pupil shift positions are computed over a grid of positions $\mathbf{s} \in [d_{min},d_{max}]$ that is determined by the window size and FFT algorithm. However, this analysis reveals that a correct implementation of STFT supervision would require the window $W(\mathbf{q})$ to also be used for generating projections of the Light Field following Eq.~\ref{eq:lf_refocusing_formula_incoherent}. As a result, Eq.~{\ref{eq:STFT2}} is not strictly valid unless the window function is a jinc function, the 2D polar analog of the sinc function, whose width depends on the wavelength:
\begin{equation}
w(\mathbf{r}) = \frac{d}{\lambda f} \text{jinc}\left(\frac{d \mathbf{r}}{\lambda f}\right),   
\end{equation}
so that its Fourier transform is a circular aperture:
\begin{equation}
W(\mathbf{q}) = A\left(\mathbf{q}, \frac{d}{\lambda f}\right)  .
\end{equation}
Focal stack and STFT supervision are, therefore, specific cases of supervising with random pupil positions, and, as a result, they overfit to specific pupil conditions, which we show does not generalize well to arbitrary viewing conditions, see in Figs.~\ref{fig:results_statistics_deepfocus} and ~\ref{fig:results:simulation_all}.     
Furthermore, because Focal Stack supervision is a special case of supervising with random pupil states, we introduce a new LF2FS algorithm that computes Focal Stacks on the fly for supervision by varying only the defocus parameter of pupil states and holding the rest constant. 
\vspace{-2mm}
\label{sec:implementation}
\subsection{Implementation}
\textit{Experimental Setup.}
We employ a conventional setup for Near-Eye Hologram as used in~\cite{peng2020neural}.
The only addition is a 2D translation and a motorized iris placed in the Fourier-plane of the 4f system to measure parallax and depth-of-field effects.
For exact setup details, we refer to the supplementary.

\textit{Implementation details.}
Optimization is implemented using ADAM \cite{kingma2014adam} using automatic-differentiation following Wirtinger Holography~\cite{chakravarthula2019wirtinger} with the standard $L_2$-loss used in all experiments.
Further implementation details on our and the comparison method STFT~\cite{choi2022time} are found in the supplementary.

\section{Results}
\label{sec:assessment}
The proposed method aims to find an optimal hologram that is optimized for the best average image quality over a four-dimensional pupil state-space, which includes diameter, location, and focus.
For all the simulations and experimental results in the paper, the pupil parameters are chosen such that the defocus range is $z_{min}=0 mm$, $z_{max}=15 mm$ and the aperture range is $d_{min}=2 mm$, $d_{max}=20 mm$.   
This is slightly smaller than the smallest eye-box size ($22mm$) that is produced by the blue channel ($440nm$) under the $400mm$ focal length that was employed in the setup with an $8um$ SLM pitch.
%
%
%
%
%
%

\textit{Comparison to Smooth-Phase Holograms.}
Our paper focuses on algorithms that produce random-phase holograms such that the maximal possible eyebox is filled.
Such holograms inherently exhibit speckle and are further harder to control due to physical effects such as field-fringing~\cite{moser2019model,schroff2023accurate}.
We expect that future research on how to combine \cite{peng2020neural}, with random phase holograms.
Recent holographic literature such as algorithms~\cite{maimone2017holographic,shi2021towards,choi2021neural} show best-in-class image quality, but rely heavily on smooth-phase to achieve speckle-reduction, which leads to a localized eye-box.
As these holograms are unlikely to achieve real 3D holography~\cite{kim2022accommodative}, we do not compare against them in our main manuscript.
However, we have added further justification and discussion of this in the supplementary materials, including a simulation using our implementation of the method from ~\cite{choi2021neural} that demonstrates how significant pupil variations are produced by methods that optimize for smooth-phase holograms.

\textit{Quantitative Validation (Simulation).}
To obtain quantitative results, we optimize our method on 85 different Light Fields from the DeepFocus~\cite{xiao2018deepfocus} dataset.
This dataset includes a variety of scenes with randomly sampled objects and textures.
Figure~\ref{fig:results_statistics_deepfocus} presents a comparison of our method with Focal Stack and STFT-supervision~\cite{choi2022time}.
Our LF2FS Focal Stack supervision method is similar to~\cite{choi2021neural,kavakli2022realistic}, however there is one significant difference:
We compute our focal stack directly from the Light Field to ensure physically correct defocus cues.
Layered approaches with inaccurate synthetic defocus, create a mismatch between projection operators which will introduce incorrect depth cues. 
%
%
%
Our findings validate that the proposed algorithm fares favorably for random pupil states other than those for which STFT-supervision and LF2FS are overfitted for. Our SLFH method produces the \emph{best-worse case performance for all pupil states}, as the best average case performance for random pupil states.
Furthermore, the variance is significantly reduced for our approaches.
This verifies that our algorithm works better for a large range of viewing conditions.

\textit{Experimental Results (Deep Focus).}
Experimental results from our prototype experimental holography setup are shown for the Deep Focus dataset in Fig~\ref{fig:teaser}.
The images are captured with a pupil diameter of $8mm$ at positions $0mm$ \textit{(back-focus)} and $12mm$~\textit{front-focus}.
In addition we extracted epipolar images by capturing a 1D-trajectory from left to right using again an $8mm$-sized pupil for both front and back focus.
Additionally, we encourage the reader to view the videos found in the Supplementary Material as differences between the result are best visible in animation. 
Our SLFH method produces fewer color artifacts than either STFT or Focal Stack supervision.

\begin{figure*}
    \centering
    \includegraphics[width=\linewidth]{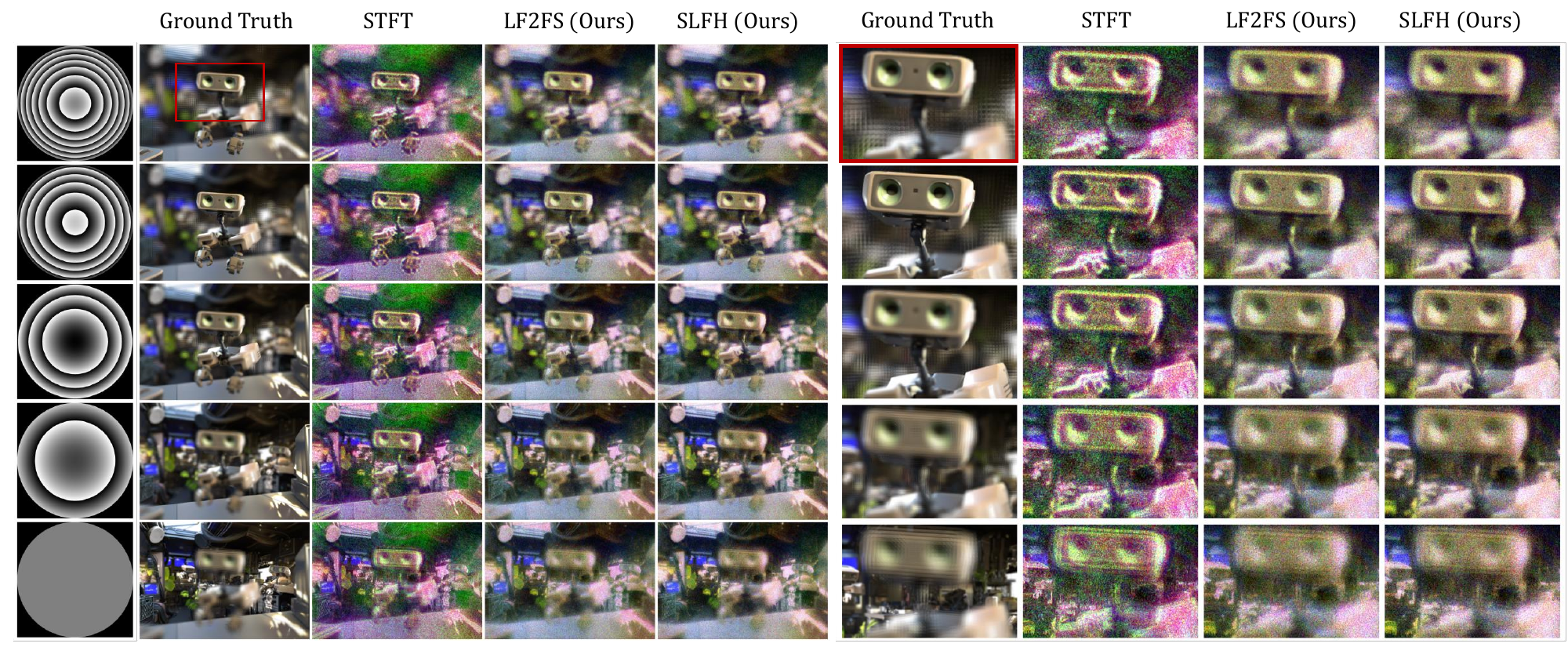}
    \caption{
       \textbf{Experiment:} \textit{ Focal-stack evaluation:}
       We show experimental results for SOTA algorithms evaluated with a wide open aperture and defocus varying from $z \in [0 mm, 13 mm]$. The STFT training method produces significant color artifacts, while our new SLFH method and our implementation of Focal Stack supervision (LF2FS) produce high quality results with similar performance.   
    }
    \label{fig:results:focal_stack_experiments}
\end{figure*}

\begin{figure*}
    \centering
    \includegraphics[width=\linewidth]{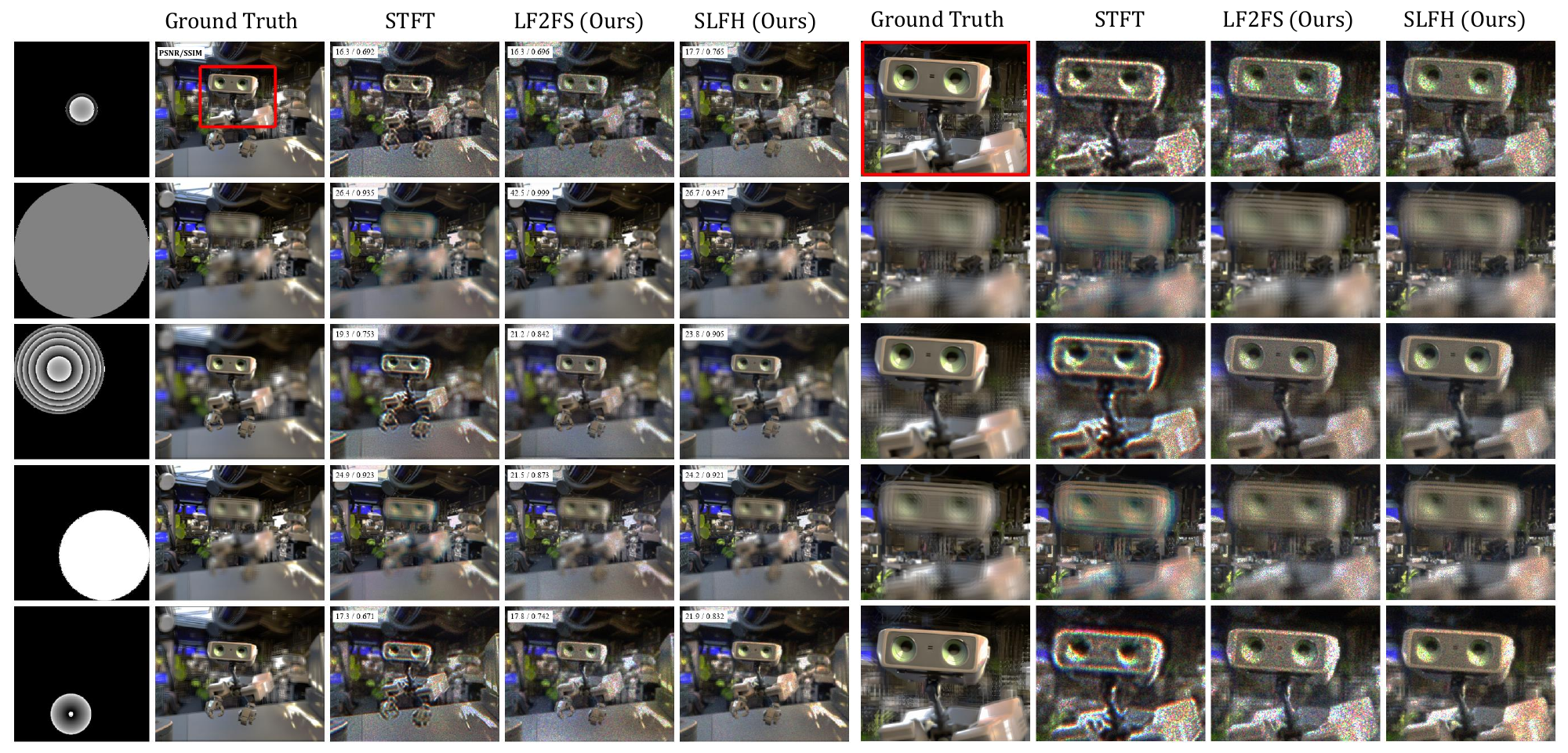}
    \caption{
       \textbf{Simulation:} \textit{ Comparison with state-of-the art:}
       We compare our proposed method with STFT-lightfield~\cite{choi2022time} and our novel implementation of Focal Stack supervision (LF2FS).
       SOTA algorithms perform close-to-perfect for the pupil-states they were over-fitted for, but fail to produce accurate results for samples they have not seen during optimization.
       As our proposed algorithm stochastically supervises with small to medium-sized pupils (diameter was 10\% to 40\% of the eyebox), we achieve overall superior image fidelity for arbitrary pupil-states.
       We evaluate the performance by testing with a set of random pupil states shown in the first column.
       Our SLFH algorithm produces significantly less noise than both LF2FS and STFT algorithms, as can be seen by PSNR/SSIM numbers in the left hand corner as well as visual comparisons.
    }
    \label{fig:results:simulation_all}
\end{figure*}

\textit{Experimental and Simulation Results (Robot Scene).}
We further evaluate our method on a blender scene in both simulation and experiments.
For experimental results, we evaluated the hologram with a focal stack trajectory as shown in Fig.~\ref{fig:results:focal_stack_experiments}.
For this, we use five centered focus positions with a large aperture (80\% of eyebox) and sample the volume in equidistant steps from $0mm$ to $12mm$.
We further evaluate the hologram in \textit{simulation} in Fig.~\ref{fig:results:simulation_all} for five randomly chosen pupil positions.
The results validate that our stochastic pupil sampling algorithm can optimize holograms with correct parallax information for large variety of pupil states.    

\begin{figure}
     \centering
    \includegraphics[width=\linewidth]{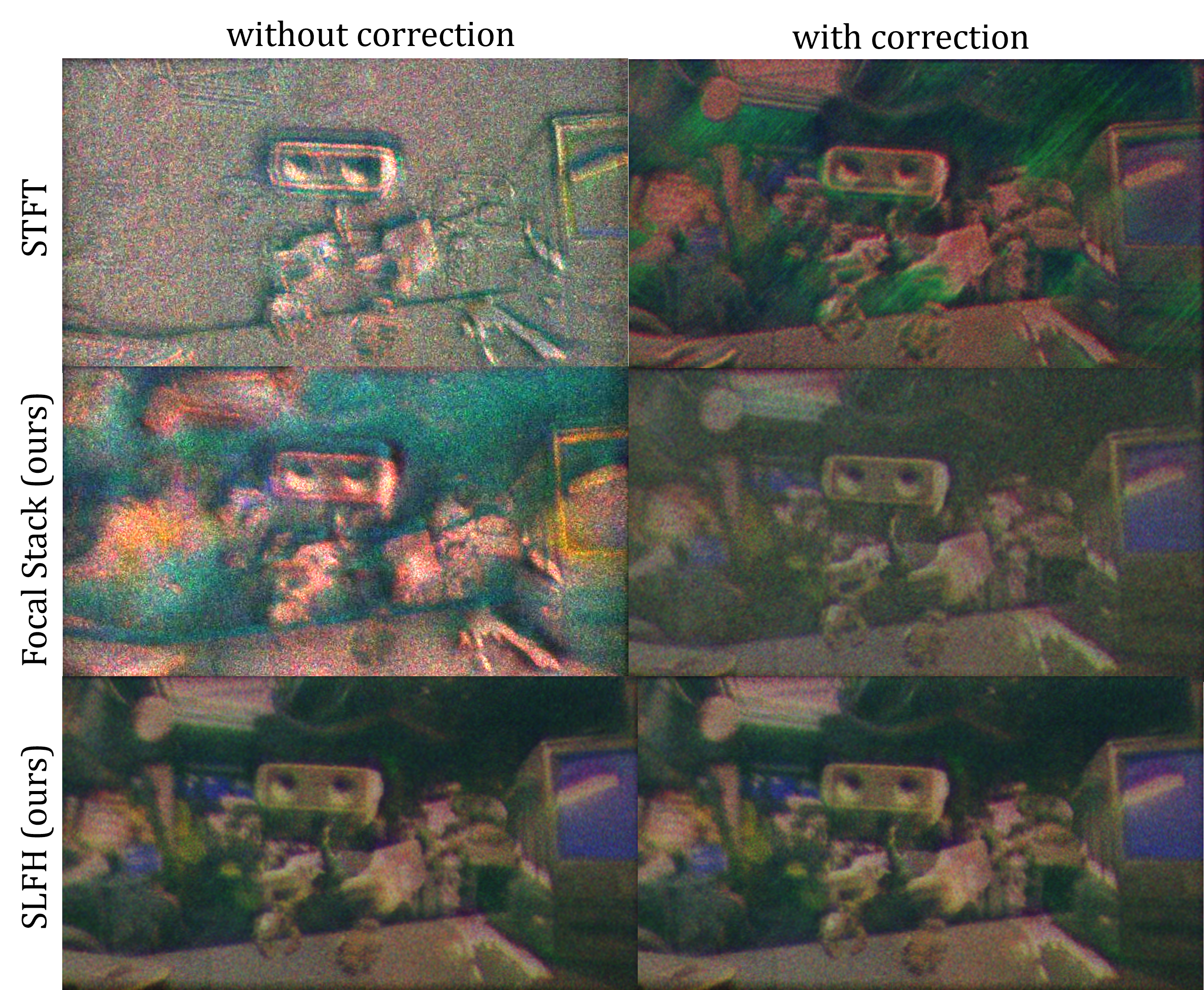}   
     \caption{\textbf{Artefacts in competing methods with and without no-care-areas at the edge of the eye-box (Experiment).}
     Plain STFT/Focal-Stack cannot produce holograms at close propagation distances (left).
     Only by introducing a bandpass filter during propagation (right), we can compute holograms for those.
     Even with correction, at the edge of the eye-box, STFT and FS strong artifacts streak artifacts or noisy images, respectively.
    }
     \label{fig:experiments_artefacts_without_and_with}
\end{figure}

\textit{Artifacts at the edge of the eye-box}
In the supplementary video, we show the parallax that is created by one hologram when we sample the eyebox from left to right.
In those, one can see that STFT is doing particularly poor at the edges of the eyebox.
This is because in order to make small propagation distances work, we had to introduce an additional bandpass filter for STFT/LF2FS, which allow some energy to diffract into no-care-areas.
Without this adaption, STFT cannot form meaningful images as shown in ~\ref{fig:experiments_artefacts_without_and_with}.
Our proposed SLFH algorithm doesn't suffer from this and the complete eyebox can be used during optimization.

\begin{figure*}
    \centering
    \includegraphics[width=\linewidth]{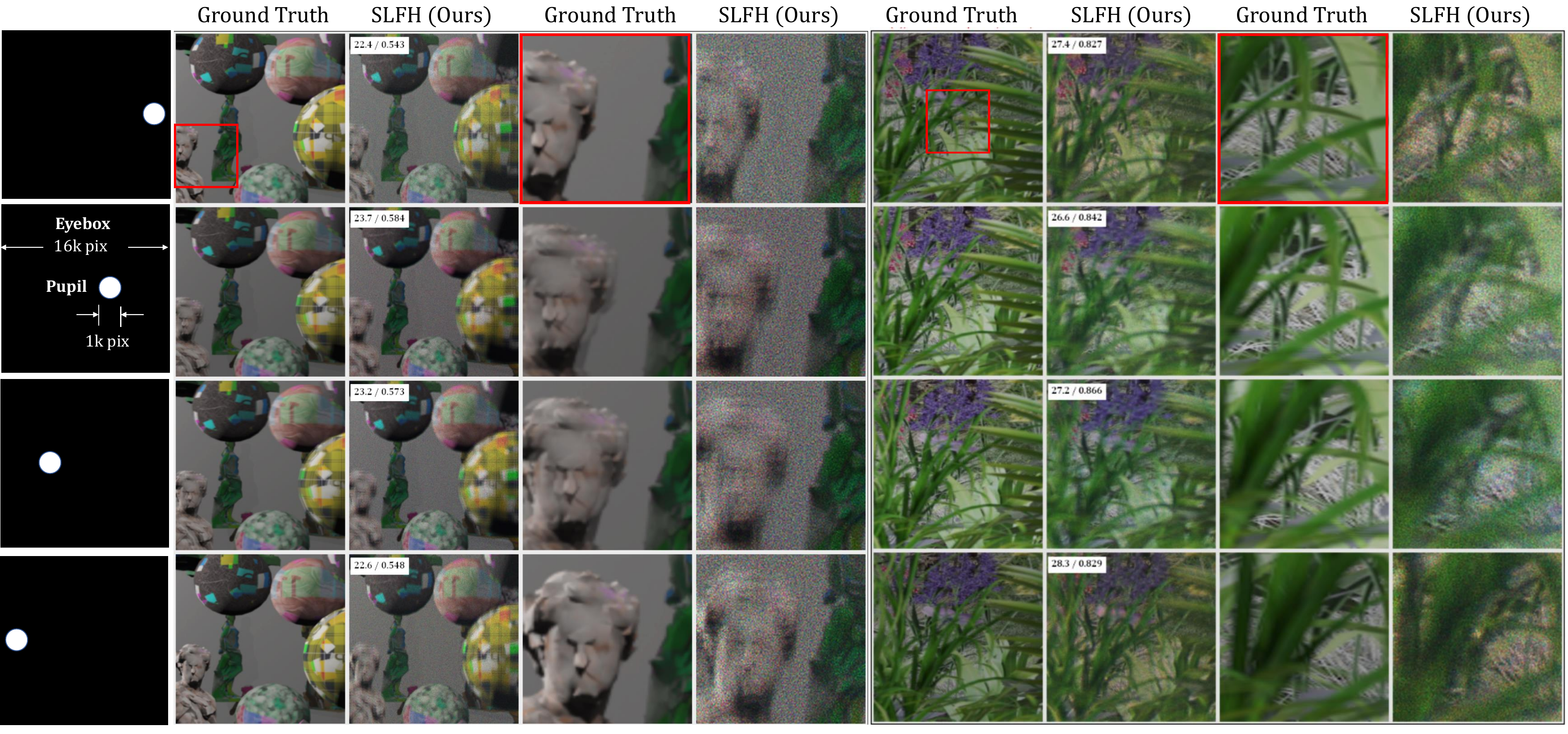}
    \caption{
       \textbf{Simulation:} \textit{Large eyebox results:}
       Here we show the benefit of our SLFH approach for optimizing over a very large eyebox in a Fourier CGH configuration. The hologram is placed in the pupil of the imaging system, and is 16k x 16k pixels, which is too large to optimize for using conventional Fourier CGH algorithms. However, our method can efficiently optimize holograms with large Space-Bandwidth Product (SBP). The figure demonstrates that larger SBP produces high quality results with significant parallax.     
    }
    \label{fig:results:SBP}
\end{figure*}

\textit{Large SBP ($16k$ x $16k$) Holograms in Far-Field Configuration.}
Current state-of-the-art displays are constrained by the etendue, which is determined by the product of the number of pixels and their pixel-pitch, also known as the Space Bandwidth Product (SBP).
As a result, there is a trade-off between parallax over a larger eye box and resolution/defocus.
This is a practical limitation for our prototype constrained by current technology, but it is not a fundamental limitation to our proposed algorithm.
With advances in technology, we can expect improvements in display technology as well as system design~\cite{kuo2020high,monin2022exponentially}
As the system etendue increases, this results in need to perform large matrix multiplications (FFTs), which can quickly exhaust available GPU memory.
%
This problem can be mitigated by optimizing for Far-Field Holograms where the angular spectrum corresponds to the field created by the SLM.
%
With a small change in the forward model and the same loss-function, we can directly supervise a large SBP hologram by stochastically sampling pupils over the Fourier-domain.
Our pupil-aware Far-Field approach is an extension to ~\cite{monin2022analyzing,monin2022exponentially} who discuss iterative algorithms in Far-Field configuration with either multi-layer or a light field loss similar to STFT~\cite{choi2022time} with a fixed pupil grid.
These works limit their discussion to only a fixed set of pupils, while we outline how Far-fields holograms with photometric consistency over the whole pupil space. 
We show an example of such a simulation for a \emph{$16k$ x $16k$ hologram} in Fig.~\ref{fig:results:SBP}.
\vspace{-2mm}

\begin{figure*}
    \centering
    \includegraphics[width=\linewidth]{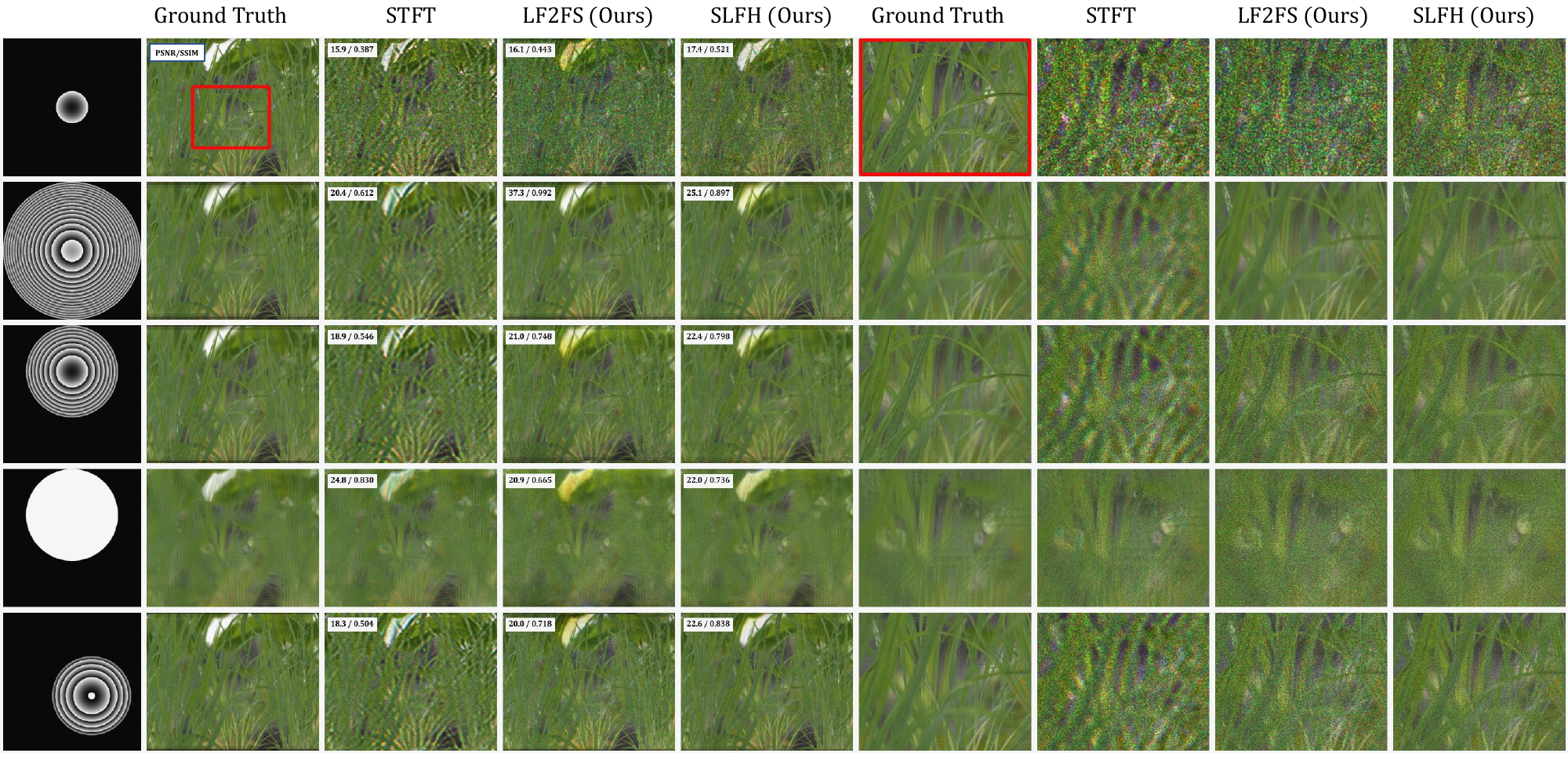}
    \caption{
       \textbf{Simulation:} \textit{ Dense occlusions:} We evaluate the performance of SOTA CGH algorithms on a scene with dense occluders consisting of a cat hiding behind dense blades of grass. We evaluate the performance by testing with a set of random pupil states shown in the first column. Our algorithm produces significantly less noise than both LF2FS and STFT algorithms, as can be seen by PSNR/SSIM numbers in the left hand corner as well as visual comparisons.
    }
    \label{fig:results:grass_scene}
\end{figure*}
%
%

\section{Discussion}
\label{sec:discussion}

We propose investigating the issue of CGH-optimization from a machine-learning perspective, specifically with regard to overfitting.
Given that the capacity of a hologram is constrained by the available SBP, there is no single, optimal solution.
In the context of our proposed approach of stochastic light-field holography optimization, the objective is to identify a hologram that yields the highest average image quality across the entire possible space of pupil states.
Our findings indicate that, while overall correct vision cues are maintained across the entire state-space, performance is inferior when evaluated specifically at the pupil states for which other algorithms were optimized.
For instance, a focal-stock loss approach may achieve near-perfect reconstruction \emph{at the depths for which it was optimized for}, but fails to accurately reproduce parallax and depth-of-field effects.
Similarly, the STFT-method demonstrates superior performance for given pupil positions around its stand-off distance, but experiences a significant decline in image quality when evaluated at larger propagation distances.

\textit{Relation to Ptychography.}
We would like to note that our method has a striking resemblance with a different technique known from Scientific Imaging called X-Ray Ptychography~\cite{shedligeri2021improving}.
There, the forward model and optimization are quite similar as a complex wavefront is reconstructed from intensity-only projections sampled at different, overlapping pupil positions.
One can think of X-Ray Ptychograpy as a dual problem in imaging compared to the holographic display problem.
Furthermore, X-Ray Ptychography often deals with very high resolutions, which requires highly parallelized, distributed optimization frameworks~\cite{nashed2017distributed} that significantly accelerate computation.
Such an increase in performance could potentially enable the development of human-sized, holographic displays.
\vspace{-3mm}
\section{Future Work} 
\textit{Light-Field Representation.}
In this study, we employed a traditional representation of a Light Field and applied the shift-and-add algorithm to synthesize new views.
However, the accuracy of the Light Field representation is constrained by its angular and spatial resolution.
Due to limitations in terms of memory, we encountered challenges such as aliasing during refocusing (details in Supplementary).
There are approaches in recent Graphics research to overcome these limitations~\cite{xiao2014aliasing,xiao2018deepfocus,yu2021plenoctrees}. 
Among those are removing angular aliasing~\cite{xiao2014aliasing}, directly synthesize refocused images using neural-networks~\cite{xiao2018deepfocus}, super-resolve in angular resolution using prior information~\cite{shedligeri2021selfvi} or fast neural-radiance fields~\cite{yu2021plenoctrees} storing a highly compressed light-field.

\textit{Speckle Reduction.}
When compared to our simulation, the image quality of our experimental results is significantly lower.
This is due to unmodeled physical effects such as cross-talk in the LCOS-SLMs.
Although camera-in-the-loop methods have been proposed, physically accurate calibration with full-random phase holograms is a novel topic~\cite{jang2022waveguide} and further research is needed.

Notably, speckle remain a fundamental challenge to holographic Light Field displays.
We use time averaging to reduce speckle artifacts, and implemented our algorithms at video rates with new high speed SLMs is a promising direction for future work.
Likewise, exploring other methods of speckle reduction~\cite{peng2021speckle} via incoherent averaging using partially coherent illumination~\cite{peng2021speckle} or subjective speckle optimization~\cite{chakravarthula2021gaze} is also a promising avenue of exploration.   

\textit{Temporal Multiplexing.}
Recent papers propose to use temporal multiplexing to either do simultaneous color~\cite{markley2023simultaneous} or reduce the overall required framerate~{\cite{kavakli2023holohdr} using joint optimization techniques.
future work could investigate how such approaches can be applied to our SLFH-algorithm to further reduce the number of temporal frames required to suppress speckle noise.

\textit{Etendue Expanders.}
Etendue for holographic displays remains impractically small due to the limited pixel count of commercially available SLMs. A further promising direction for future work is to explore our stochastic pupil-sampling cgh algorithms together with etendue expanders~\cite{baek2021neural} that can help meet AR/VR requirements for combined eyebox size and FoV~\cite{xia2020towards}.

\section{Conclusion}
\label{sec:conclusion}
We introduce a new optimization method for generating light-field holograms using stochastic pupil-sampling within a Gradient-Descent type optimization.
Simulation results demonstrating the effectiveness of the proposed method in near-eye display and large etendue setups in far-field configurations were presented.
Experimental results validate that our algorithm is able to provide accurate depth cues such as parallax and defocus under a large variety of different pupil states. All our findings confirm that the proposed method produces the best-worse case performance for all pupil states, as well as the best average case performance for random pupil states -- corresponding to average viewing conditions.


\bibliographystyle{IEEEtran}
\bibliography{references}

\ifpeerreview \else



\begin{IEEEbiographynophoto}{Florian Schiffers}~is a Ph.D. candidate at Northwestern University and has undertaken the research of this paper during a research internship at Reality Labs Research, Meta. 
He received dual M.Sc. degrees, in Physics and Advanced Optical Technologies, from FAU Erlangen in Germany in 2017. Prior to his ongoing doctoral study, Schiffers held a research assistant role at Siemens Healthineers.
\end{IEEEbiographynophoto}
\vspace{-10mm}
\begin{IEEEbiographynophoto}{Praneeth Chakravarthula}
is a research scholar at Princeton University in the Princeton Computational Imaging Lab.
~
He obtained his Ph.D. from UNC Chapel Hill on novel near-eye displays for virtual and augmented reality.
His research interests lie at the intersection of optics, perception, and graphics.
Prior to joining UNC, he obtained my B.Tech and M.Tech degrees in Electrical Engineering from IIT Madras. 
\end{IEEEbiographynophoto}
\vspace{-10mm}
\begin{IEEEbiographynophoto}{Nathan Matsuda}
is a research scientist at Meta developing computational camera and display systems that support immersive visual experiences in VR.
\end{IEEEbiographynophoto}
\vspace{-10mm}
\begin{IEEEbiographynophoto}{Grace Kuo}
is a research scientist at Reality Labs Research, Meta working on the joint design of optical hardware and algorithms for imaging and display systems. She earned her PhD from the Department of Electrical Engineering and Computer Sciences at UC Berkeley, advised by Dr. Laura Waller and Dr. Ren Ng.
\end{IEEEbiographynophoto}
\vspace{-10mm}
\begin{IEEEbiographynophoto}{Ethan Tseng}
 is currently a Ph.D. candidate in Computer Science at Princeton University, supervised by Prof. Felix Heide. He received his B.S. in Electrical and Computer Engineering from Carnegie Mellon University.
\end{IEEEbiographynophoto}
\vspace{-10mm}
\begin{IEEEbiographynophoto}{Douglas Lanman}
is the Senior Director of Display Systems Research at Reality Labs Research, Meta, where he leads investigations into advanced display and imaging technologies.
His prior research focuses on head-mounted displays, glasses-free 3D displays, light field cameras, and active illumination for 3D reconstruction and interaction.
Douglas holds a B.S. in Applied Physics from Caltech, and M.S. and Ph.D. degrees in Electrical Engineering from Brown University.
\end{IEEEbiographynophoto}
\vspace{-10mm}
\begin{IEEEbiographynophoto}{Felix Heide} is an assistant professor at Princeton University, where he leads the Computational Imaging Lab.
He obtained his BS and MS from the University of Siegen and his PhD from the University of British Columbia, where his doctoral dissertation won the Alain Fournier PhD Dissertation Award and the SIGGRAPH Outstanding Doctoral Dissertation Award.
He completed his postdoc at Stanford University and has been at Princeton since 2020. 
\end{IEEEbiographynophoto}
%
\vspace{-10mm}
\begin{IEEEbiographynophoto}{Oliver S. Cossairt}
 is a research scientist at Reality Labs Research, Meta and an Associate Professor in the Electrical Engineering and Computer Science Department at Northwestern University. He holds an M.S. from the MIT Media Lab, and a Ph.D. in Computer Science from Columbia University. With a background as an Optical and Software Engineer at Actuality Systems, Oliver has received several accolades, including the NSF Graduate Research Fellowship, ICCP Best Paper and Honorable Mention, and an NSF CAREER Award.
\end{IEEEbiographynophoto}
\vfill

\fi

\end{document}